\documentclass[sigconf]{acmart}
\AtBeginDocument{%
  }

\setcopyright{acmlicensed}
\copyrightyear{2025}
\acmYear{2025}

\begin{document}

\title{Perception Graph for Cognitive Attack Reasoning in Augmented Reality}


\author{Rongqian Chen}
\affiliation{%
  \institution{George Washington University}
  \state{Washington D.C.}
  \country{USA}
}
\email{rongqianc@gwu.edu}

\author{Shu Hong}
\affiliation{%
  \institution{George Washington University}
  \state{Washington D.C.}
  \country{USA}
}
\email{shu.hong@gwu.edu}

\author{Rifatul Islam}
\affiliation{%
  \institution{Kennesaw State University}
  \city{Kennesaw}
  \state{GA}
  \country{USA}
}
\email{rislam11@kennesaw.edu}

\author{Mahdi Imani}
\affiliation{%
  \institution{Northeastern University}
  \city{Boston}
  \state{MA}
  \country{USA}
}
\email{m.imani@northeastern.edu}

\author{G. Gary Tan}
\affiliation{%
  \institution{Pennsylvania State University}
  \city{University Park}
  \state{PA}
  \country{USA}
}
\email{gtan@psu.edu}

\author{Tian Lan}
\affiliation{%
  \institution{George Washington University}
  \state{Washington D.C.}
  \country{USA}
  }
\email{tlan@gwu.edu}


\begin{abstract}
  Augmented reality (AR) systems are increasingly deployed in tactical environments, but their reliance on seamless human-computer interaction makes them vulnerable to cognitive attacks that manipulate a user's perception and severely compromise user decision-making. To address this challenge, we introduce the Perception Graph, a novel model designed to reason about human perception within these systems. Our model operates by first mimicking the human process of interpreting key information from an MR environment and then representing the outcomes using a semantically meaningful structure.  We demonstrate how the model can compute a quantitative score that reflects the level of perception distortion, providing a robust and measurable method for detecting and analyzing the effects of such cognitive attacks.
\end{abstract}

\begin{teaserfigure}
  \includegraphics[width=\textwidth]{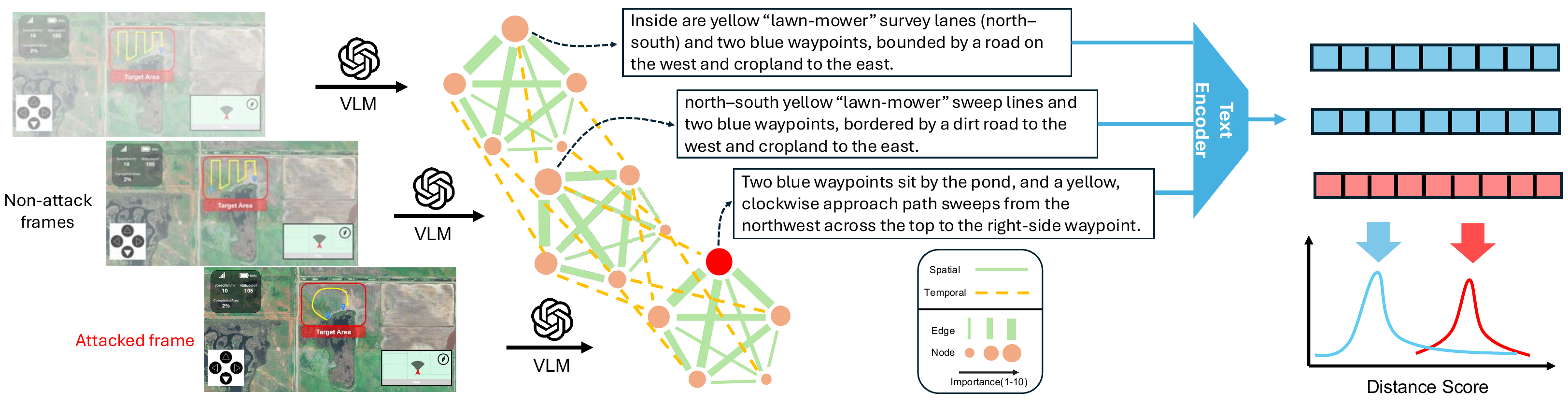}
  \caption{Perception Graph overview — cognitive knowledge is spatially and temporally stored, encoded, and reasoned upon. }
  \label{fig:system overview}
\end{teaserfigure}


\maketitle


\begin{figure*}[tb]
    \centering
    \includegraphics[width=\linewidth]{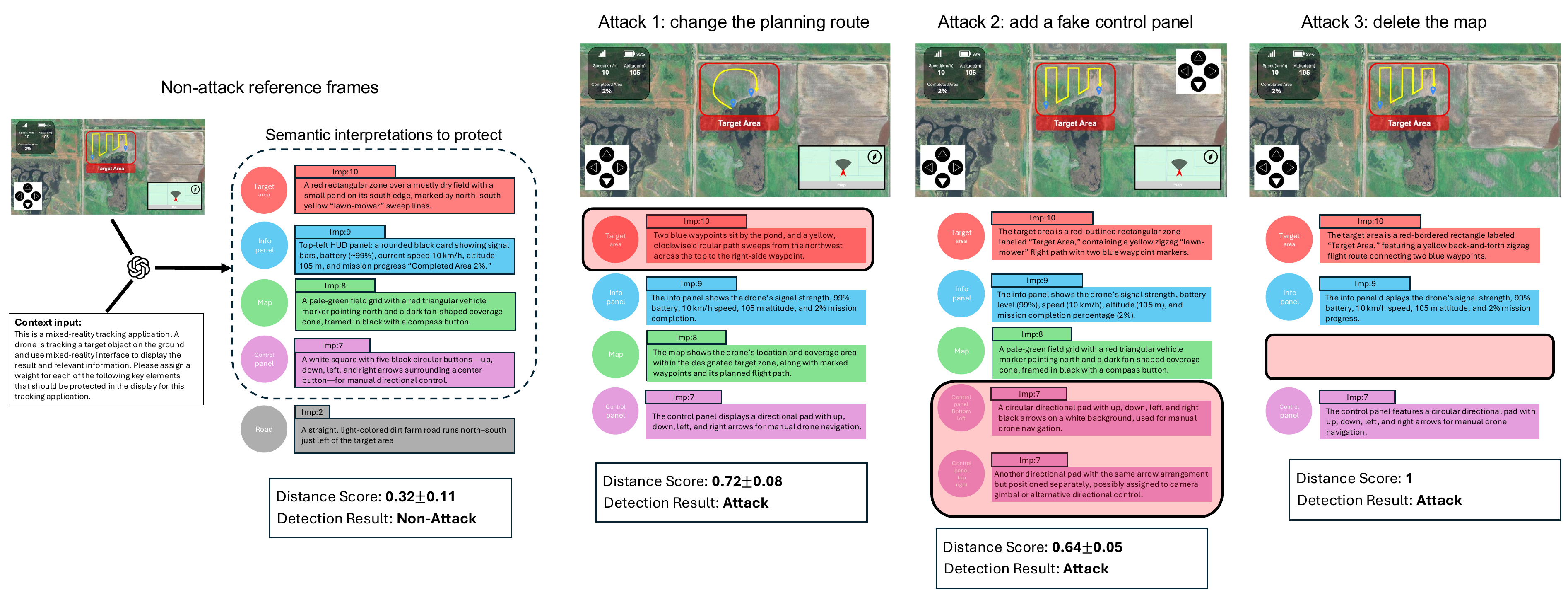}
    \caption{Attack reasoning examples in an agricultural drone scene — alterations in Perception Graph information or structural changes result in higher distance scores.}
    \label{fig:result}
\end{figure*}

\section{Introduction}

In augmented reality (AR) systems, the seamless blending of digital and physical worlds introduces new vulnerabilities to cognitive attacks that manipulate human perception. Such attacks—like inserting fake objects or removing real ones—can severely impair a user’s ability to interpret critical information. Traditional computer vision models, which typically operate at the pixel level, often fail to detect these semantically meaningful alterations. Supervised learning approaches also face limitations, as they require large amounts of training data, which is difficult to collect in real AR environments and across diverse attack patterns, especially in safety-critical scenarios.

To address these limitations, we developed the Perception Graph model, a novel approach for reasoning about human perception in a few shots in AR systems. Our model uses pre-trained vision language models (VLMs) to mimic human interpretation and understanding~\cite{chen2025neurosymbolic,liu2023auxiliary,liu2023iterative,liu2024dvlo,liu2025mamba4d,li2024crowdsensing,liu2025auxiliary,erbayat2025lamps,liu2025topolidm,guo2023advantage,jiang2022intelligent,ravari2024adversarial,tang2023edge,wu2025probabilistic_ismar,mei2024projection,xu2023cnn,yu2025optimizingpromptsequencesusing,zhao2024balf,liu2024learning, xiu2025viddar,yu2025look,xiu2025detecting,liu2024safety,fang2023implementing,fang2024coordinate,fang2024learning}. It converts AR visual inputs into a compact, semantically rich representation, which takes contextual factors into account. This structured approach allows us to not only identify changes in the perception space but also to compute a quantitative distortion score, reflecting the impact of cognitive attacks on human interpretation of information. Our model offers human-like understanding compared to traditional black-box processes, providing a robust, interpretable, and trustworthy foundation for perception security in mission-critical applications.

\section{Methodology}

Our Perception Graph model is designed to detect cognitive attacks in AR environments through a two-phase process.
\paragraph{Graph Construction Phase}
As illustrated in Fig.\ref{fig:system overview} and Fig.\ref{fig:result}, the construction phase begins by generating reference graphs that encode the ground-truth semantic interpretation of a scene. These graphs are derived from the semantic outputs of Vision–Language Models (VLMs). While VLMs capture the underlying meaning of objects and relationships, their natural language descriptions often vary in word choice and phrasing. To resolve this variability, each description is passed through a text encoder, which projects it into a latent embedding space. In this space, semantic meaning is represented by the direction of the embedding vector, and similarities between descriptions can be consistently measured with cosine similarity.

In addition to encoding meaning, the construction phase also generate contextual weights for each object. These weights quantify the object’s relative importance in the scene, allowing the model to focus protection efforts on critical objects. For example, traffic signs, hazard warnings, or navigation markers—while treating less relevant objects with lower priority. This selective emphasis ensures that detection resources are concentrated where cognitive attacks would cause the greatest harm.

\paragraph{Detection phase}

During the detection phase, the model processes new AR frames to generate a perception graph and aligns it with the stored reference graphs. Scene objects are represented as nodes carrying both semantic embeddings and contextual weights. Semantic changes—such as addition, removal, or modification of nodes—are detected by comparing embeddings of corresponding nodes. To quantify differences, we define a distance function: $Distance = \sqrt{1 - Sim(E_1, E_2)}$, where $Sim(E_1, E_2)$ denotes the cosine similarity between embeddings $E_1$ and $E_2$. Smaller distances indicate semantic consistency, while larger distances reveal deviations. A distance of 1 corresponds to a missing node (i.e., no semantic match), and distances exceeding a defined threshold trigger a potential cognitive attack alert.

\section{Demonstration}
\begin{table}[t]
\centering
\caption{Attack detection based on distance scores and Z-scores relative to the normal distribution ($\mu = 0.32$, $\sigma = 0.11$).}
\label{tab:attack_zscore}
\begin{tabular}{lccc}
\toprule
\textbf{Attack Type} & \textbf{Distance} & \textbf{Z-score} & \textbf{Detection}\\
\midrule
Route Modification   & $0.72 \pm 0.08$ & $3.6\sigma$ & Attack \\
Fake Control Panel   & $0.64 \pm 0.05$ & $2.9\sigma$ & Attack \\
Map Deletion         & $1.00$          & $6.2\sigma$ & Attack \\
\bottomrule
\end{tabular}
\end{table}

In Fig.~\ref{fig:result}, we show three representative cognitive attacks and their impact on graph distance scores over 10 reference frames. Under normal conditions, distances follow a distribution with mean $\mu = 0.32$ and standard deviation $\sigma = 0.11$, capturing natural variation in VLM-generated embeddings.  

To quantify deviations, we compute the \emph{Z-score} for each observed distance $d$:  
\begin{equation}
Z = \frac{d - \mu}{\sigma}.
\end{equation}  

Table~\ref{tab:attack_zscore} reports the results. Route modification yields $3.6\sigma$, fake control panel $2.9\sigma$, and map deletion $6.2\sigma$—all well outside normal variation. Frames exceeding a threshold (e.g., $Z > 2$) are flagged as potential cognitive attacks.  

This statistical mapping transforms raw distances into interpretable evidence of semantic deviation, enabling robust detection of AR cognitive attacks.

\bibliographystyle{ACM-Reference-Format}
\bibliography{references}


\begin{thebibliography}{26}


\ifx \showCODEN    \undefined \def \showCODEN     #1{\unskip}     \fi
\ifx \showDOI      \undefined \def \showDOI       #1{#1}\fi
\ifx \showISBNx    \undefined \def \showISBNx     #1{\unskip}     \fi
\ifx \showISBNxiii \undefined \def \showISBNxiii  #1{\unskip}     \fi
\ifx \showISSN     \undefined \def \showISSN      #1{\unskip}     \fi
\ifx \showLCCN     \undefined \def \showLCCN      #1{\unskip}     \fi
\ifx \shownote     \undefined \def \shownote      #1{#1}          \fi
\ifx \showarticletitle \undefined \def \showarticletitle #1{#1}   \fi
\ifx \showURL      \undefined \def \showURL       {\relax}        \fi
\providecommand\bibfield[2]{#2}
\providecommand\bibinfo[2]{#2}
\providecommand\natexlab[1]{#1}
\providecommand\showeprint[2][]{arXiv:#2}

\bibitem[Chen et~al\mbox{.}(2025)]%
        {chen2025neurosymbolic}
\bibfield{author}{\bibinfo{person}{Rongqian Chen}, \bibinfo{person}{Allison Andreyev}, \bibinfo{person}{Yanming Xiu}, \bibinfo{person}{Mahdi Imani}, \bibinfo{person}{Bin Li}, \bibinfo{person}{Maria Gorlatova}, \bibinfo{person}{Gang Tan}, {and} \bibinfo{person}{Tian Lan}.} \bibinfo{year}{2025}\natexlab{}.
\newblock \showarticletitle{A Neurosymbolic Framework for Interpretable Cognitive Attack Detection in Augmented Reality}.
\newblock \bibinfo{journal}{\emph{arXiv preprint arXiv:2508.09185}} (\bibinfo{year}{2025}).
\newblock


\bibitem[Erbayat et~al\mbox{.}(2025)]%
        {erbayat2025lamps}
\bibfield{author}{\bibinfo{person}{Egemen Erbayat}, \bibinfo{person}{Yongsheng Mei}, \bibinfo{person}{Gina Adam}, \bibinfo{person}{Suresh Subramaniam}, \bibinfo{person}{Sean Coffey}, \bibinfo{person}{Nathaniel~D Bastian}, {and} \bibinfo{person}{Tian Lan}.} \bibinfo{year}{2025}\natexlab{}.
\newblock \showarticletitle{LAMPS: A Learning-based Mobility Planning via Posterior$\backslash$$\backslash$State Inference using Gaussian Cox Process Models}. In \bibinfo{booktitle}{\emph{AAAI 2025 Workshop on Artificial Intelligence for Wireless Communications and Networking (AI4WCN)}}.
\newblock


\bibitem[Fang and Lan(2024)]%
        {fang2024learning}
\bibfield{author}{\bibinfo{person}{Zeyu Fang} {and} \bibinfo{person}{Tian Lan}.} \bibinfo{year}{2024}\natexlab{}.
\newblock \showarticletitle{Learning from random demonstrations: Offline reinforcement learning with importance-sampled diffusion models}.
\newblock \bibinfo{journal}{\emph{arXiv preprint arXiv:2405.19878}} (\bibinfo{year}{2024}).
\newblock


\bibitem[Fang et~al\mbox{.}(2024)]%
        {fang2024coordinate}
\bibfield{author}{\bibinfo{person}{Zeyu Fang}, \bibinfo{person}{Jian Zhao}, \bibinfo{person}{Mingyu Yang}, \bibinfo{person}{Zhenbo Lu}, \bibinfo{person}{Wengang Zhou}, {and} \bibinfo{person}{Houqiang Li}.} \bibinfo{year}{2024}\natexlab{}.
\newblock \showarticletitle{Coordinate-aligned multi-camera collaboration for active multi-object tracking}.
\newblock \bibinfo{journal}{\emph{Multimedia Systems}} \bibinfo{volume}{30}, \bibinfo{number}{4} (\bibinfo{year}{2024}), \bibinfo{pages}{221}.
\newblock


\bibitem[Fang et~al\mbox{.}(2023)]%
        {fang2023implementing}
\bibfield{author}{\bibinfo{person}{Zeyu Fang}, \bibinfo{person}{Jian Zhao}, \bibinfo{person}{Wengang Zhou}, {and} \bibinfo{person}{Houqiang Li}.} \bibinfo{year}{2023}\natexlab{}.
\newblock \showarticletitle{Implementing first-person shooter game AI in WILD-SCAV with rule-enhanced deep reinforcement learning}. In \bibinfo{booktitle}{\emph{2023 IEEE Conference on Games (CoG)}}. IEEE, \bibinfo{pages}{1--8}.
\newblock


\bibitem[Guo et~al\mbox{.}(2023)]%
        {guo2023advantage}
\bibfield{author}{\bibinfo{person}{Muzhe Guo}, \bibinfo{person}{Feixu Yu}, \bibinfo{person}{Tian Lan}, {and} \bibinfo{person}{Fang Jin}.} \bibinfo{year}{2023}\natexlab{}.
\newblock \showarticletitle{Advantage Actor-Critic with Reasoner: Explaining the Agent's Behavior from an Exploratory Perspective}.
\newblock \bibinfo{journal}{\emph{arXiv preprint arXiv:2309.04707}} (\bibinfo{year}{2023}).
\newblock


\bibitem[Jiang et~al\mbox{.}(2022)]%
        {jiang2022intelligent}
\bibfield{author}{\bibinfo{person}{Qinting Jiang}, \bibinfo{person}{Xuanhong Zhou}, \bibinfo{person}{Ruili Wang}, \bibinfo{person}{Weiping Ding}, \bibinfo{person}{Yi Chu}, \bibinfo{person}{Sizhe Tang}, \bibinfo{person}{Xiaoyun Jia}, {and} \bibinfo{person}{Xiaolong Xu}.} \bibinfo{year}{2022}\natexlab{}.
\newblock \showarticletitle{Intelligent monitoring for infectious diseases with fuzzy systems and edge computing: A survey}.
\newblock \bibinfo{journal}{\emph{Applied Soft Computing}}  \bibinfo{volume}{123} (\bibinfo{year}{2022}), \bibinfo{pages}{108835}.
\newblock


\bibitem[Li et~al\mbox{.}(2024)]%
        {li2024crowdsensing}
\bibfield{author}{\bibinfo{person}{Zheng Li}, \bibinfo{person}{Sizhe Tang}, \bibinfo{person}{Hao Tian}, \bibinfo{person}{Haolong Xiang}, \bibinfo{person}{Xiaolong Xu}, {and} \bibinfo{person}{Wanchun Dou}.} \bibinfo{year}{2024}\natexlab{}.
\newblock \showarticletitle{A Crowdsensing Service Pricing Method in Vehicular Edge Computing}. In \bibinfo{booktitle}{\emph{2024 IEEE International Symposium on Parallel and Distributed Processing with Applications (ISPA)}}. IEEE, \bibinfo{pages}{82--89}.
\newblock


\bibitem[Liu et~al\mbox{.}(2025a)]%
        {liu2025mamba4d}
\bibfield{author}{\bibinfo{person}{Jiuming Liu}, \bibinfo{person}{Jinru Han}, \bibinfo{person}{Lihao Liu}, \bibinfo{person}{Angelica~I Aviles-Rivero}, \bibinfo{person}{Chaokang Jiang}, \bibinfo{person}{Zhe Liu}, {and} \bibinfo{person}{Hesheng Wang}.} \bibinfo{year}{2025}\natexlab{a}.
\newblock \showarticletitle{Mamba4D: Efficient 4D Point Cloud Video Understanding with Disentangled Spatial-Temporal State Space Models}. In \bibinfo{booktitle}{\emph{Proceedings of the Computer Vision and Pattern Recognition Conference}}. \bibinfo{pages}{17626--17636}.
\newblock


\bibitem[Liu et~al\mbox{.}(2025b)]%
        {liu2025topolidm}
\bibfield{author}{\bibinfo{person}{Jiuming Liu}, \bibinfo{person}{Zheng Huang}, \bibinfo{person}{Mengmeng Liu}, \bibinfo{person}{Tianchen Deng}, \bibinfo{person}{Francesco Nex}, \bibinfo{person}{Hao Cheng}, {and} \bibinfo{person}{Hesheng Wang}.} \bibinfo{year}{2025}\natexlab{b}.
\newblock \showarticletitle{TopoLiDM: Topology-Aware LiDAR Diffusion Models for Interpretable and Realistic LiDAR Point Cloud Generation}.
\newblock \bibinfo{journal}{\emph{arXiv preprint arXiv:2507.22454}} (\bibinfo{year}{2025}).
\newblock


\bibitem[Liu et~al\mbox{.}(2024c)]%
        {liu2024dvlo}
\bibfield{author}{\bibinfo{person}{Jiuming Liu}, \bibinfo{person}{Dong Zhuo}, \bibinfo{person}{Zhiheng Feng}, \bibinfo{person}{Siting Zhu}, \bibinfo{person}{Chensheng Peng}, \bibinfo{person}{Zhe Liu}, {and} \bibinfo{person}{Hesheng Wang}.} \bibinfo{year}{2024}\natexlab{c}.
\newblock \showarticletitle{Dvlo: Deep visual-lidar odometry with local-to-global feature fusion and bi-directional structure alignment}. In \bibinfo{booktitle}{\emph{European Conference on Computer Vision}}. Springer, \bibinfo{pages}{475--493}.
\newblock


\bibitem[Liu et~al\mbox{.}(2024a)]%
        {liu2024learning}
\bibfield{author}{\bibinfo{person}{Shuo Liu}, \bibinfo{person}{Zhe Huang}, \bibinfo{person}{Jun Zeng}, \bibinfo{person}{Koushil Sreenath}, {and} \bibinfo{person}{Calin~A Belta}.} \bibinfo{year}{2024}\natexlab{a}.
\newblock \showarticletitle{Learning-Enabled Iterative Convex Optimization for Safety-Critical Model Predictive Control}.
\newblock \bibinfo{journal}{\emph{arXiv preprint arXiv:2409.08300}} (\bibinfo{year}{2024}).
\newblock


\bibitem[Liu et~al\mbox{.}(2024b)]%
        {liu2024safety}
\bibfield{author}{\bibinfo{person}{Shuo Liu}, \bibinfo{person}{Yihui Mao}, {and} \bibinfo{person}{Calin~A Belta}.} \bibinfo{year}{2024}\natexlab{b}.
\newblock \showarticletitle{Safety-critical planning and control for dynamic obstacle avoidance using control barrier functions}.
\newblock \bibinfo{journal}{\emph{arXiv preprint arXiv:2403.19122}} (\bibinfo{year}{2024}).
\newblock


\bibitem[Liu et~al\mbox{.}(2023a)]%
        {liu2023auxiliary}
\bibfield{author}{\bibinfo{person}{Shuo Liu}, \bibinfo{person}{Wei Xiao}, {and} \bibinfo{person}{Calin~A Belta}.} \bibinfo{year}{2023}\natexlab{a}.
\newblock \showarticletitle{Auxiliary-variable adaptive control barrier functions for safety critical systems}. In \bibinfo{booktitle}{\emph{2023 62nd IEEE Conference on Decision and Control (CDC)}}. IEEE, \bibinfo{pages}{8602--8607}.
\newblock


\bibitem[Liu et~al\mbox{.}(2025c)]%
        {liu2025auxiliary}
\bibfield{author}{\bibinfo{person}{Shuo Liu}, \bibinfo{person}{Wei Xiao}, {and} \bibinfo{person}{Calin~A Belta}.} \bibinfo{year}{2025}\natexlab{c}.
\newblock \showarticletitle{Auxiliary-Variable Adaptive Control Barrier Functions}.
\newblock \bibinfo{journal}{\emph{arXiv preprint arXiv:2502.15026}} (\bibinfo{year}{2025}).
\newblock


\bibitem[Liu et~al\mbox{.}(2023b)]%
        {liu2023iterative}
\bibfield{author}{\bibinfo{person}{Shuo Liu}, \bibinfo{person}{Jun Zeng}, \bibinfo{person}{Koushil Sreenath}, {and} \bibinfo{person}{Calin~A Belta}.} \bibinfo{year}{2023}\natexlab{b}.
\newblock \showarticletitle{Iterative Convex Optimization for Model Predictive Control with Discrete-Time High-Order Control Barrier Functions}. In \bibinfo{booktitle}{\emph{2023 American Control Conference (ACC)}}. IEEE, \bibinfo{pages}{3368--3375}.
\newblock


\bibitem[Mei et~al\mbox{.}(2024)]%
        {mei2024projection}
\bibfield{author}{\bibinfo{person}{Yongsheng Mei}, \bibinfo{person}{Hanhan Zhou}, {and} \bibinfo{person}{Tian Lan}.} \bibinfo{year}{2024}\natexlab{}.
\newblock \showarticletitle{Projection-Optimal Monotonic Value Function Factorization in Multi-Agent Reinforcement Learning.}. In \bibinfo{booktitle}{\emph{AAMAS}}. \bibinfo{pages}{2381--2383}.
\newblock


\bibitem[Ravari et~al\mbox{.}(2024)]%
        {ravari2024adversarial}
\bibfield{author}{\bibinfo{person}{Amirhossein Ravari}, \bibinfo{person}{Guangyu Jiang}, \bibinfo{person}{Zuyuan Zhang}, \bibinfo{person}{Mahdi Imani}, \bibinfo{person}{Robert~H Thomson}, \bibinfo{person}{Aryn~A Pyke}, \bibinfo{person}{Nathaniel~D Bastian}, {and} \bibinfo{person}{Tian Lan}.} \bibinfo{year}{2024}\natexlab{}.
\newblock \showarticletitle{Adversarial inverse learning of defense policies conditioned on human factor models}. In \bibinfo{booktitle}{\emph{2024 58th Asilomar Conference on Signals, Systems, and Computers}}. IEEE, \bibinfo{pages}{188--195}.
\newblock


\bibitem[Tang et~al\mbox{.}(2023)]%
        {tang2023edge}
\bibfield{author}{\bibinfo{person}{Sizhe Tang}, \bibinfo{person}{Mengmeng Cui}, \bibinfo{person}{Lianyong Qi}, {and} \bibinfo{person}{Xiaolong Xu}.} \bibinfo{year}{2023}\natexlab{}.
\newblock \showarticletitle{Edge Intelligence with Distributed Processing of DNNs: A Survey.}
\newblock \bibinfo{journal}{\emph{CMES-Computer Modeling in Engineering \& Sciences}} \bibinfo{volume}{136}, \bibinfo{number}{1} (\bibinfo{year}{2023}).
\newblock


\bibitem[Wu et~al\mbox{.}(2025)]%
        {wu2025probabilistic_ismar}
\bibfield{author}{\bibinfo{person}{Peng Wu}, \bibinfo{person}{Nasim Ahmed}, \bibinfo{person}{Abhiram Sarma}, \bibinfo{person}{Kaiming Huang}, \bibinfo{person}{Rifatul Islam}, \bibinfo{person}{Bin Li}, \bibinfo{person}{Tian Lan}, \bibinfo{person}{Gang Tan}, {and} \bibinfo{person}{Mahdi Imani}.} \bibinfo{year}{2025}\natexlab{}.
\newblock \showarticletitle{Probabilistic Verification of Cybersickness in Virtual Reality Through Bayesian Networks}. In \bibinfo{booktitle}{\emph{Proceedings of the IEEE International Symposium on Mixed and Augmented Reality (ISMAR)}}. \bibinfo{address}{Daejeon, South Korea}.
\newblock
\newblock
\shownote{To appear}.


\bibitem[Xiu and Gorlatova(2025)]%
        {xiu2025detecting}
\bibfield{author}{\bibinfo{person}{Yanming Xiu} {and} \bibinfo{person}{Maria Gorlatova}.} \bibinfo{year}{2025}\natexlab{}.
\newblock \showarticletitle{Detecting visual information manipulation attacks in augmented reality: a multimodal semantic reasoning approach}.
\newblock \bibinfo{journal}{\emph{arXiv preprint arXiv:2507.20356}} (\bibinfo{year}{2025}).
\newblock


\bibitem[Xiu et~al\mbox{.}(2025)]%
        {xiu2025viddar}
\bibfield{author}{\bibinfo{person}{Yanming Xiu}, \bibinfo{person}{Tim Scargill}, {and} \bibinfo{person}{Maria Gorlatova}.} \bibinfo{year}{2025}\natexlab{}.
\newblock \showarticletitle{ViDDAR: Vision language model-based task-detrimental content detection for augmented reality}.
\newblock \bibinfo{journal}{\emph{IEEE transactions on visualization and computer graphics}} (\bibinfo{year}{2025}).
\newblock


\bibitem[Xu et~al\mbox{.}(2023)]%
        {xu2023cnn}
\bibfield{author}{\bibinfo{person}{Xiaolong Xu}, \bibinfo{person}{Sizhe Tang}, \bibinfo{person}{Lianyong Qi}, \bibinfo{person}{Xiaokang Zhou}, \bibinfo{person}{Fei Dai}, {and} \bibinfo{person}{Wanchun Dou}.} \bibinfo{year}{2023}\natexlab{}.
\newblock \showarticletitle{Cnn partitioning and offloading for vehicular edge networks in web3}.
\newblock \bibinfo{journal}{\emph{IEEE Communications Magazine}} \bibinfo{volume}{61}, \bibinfo{number}{8} (\bibinfo{year}{2023}), \bibinfo{pages}{36--42}.
\newblock


\bibitem[Yu et~al\mbox{.}(2025)]%
        {yu2025optimizingpromptsequencesusing}
\bibfield{author}{\bibinfo{person}{Fei~Xu Yu}, \bibinfo{person}{Gina Adam}, \bibinfo{person}{Nathaniel~D. Bastian}, {and} \bibinfo{person}{Tian Lan}.} \bibinfo{year}{2025}\natexlab{}.
\newblock \bibinfo{title}{Optimizing Prompt Sequences using Monte Carlo Tree Search for LLM-Based Optimization}.
\newblock
\newblock
\showeprint[arxiv]{2508.05995}~[cs.LG]
\urldef\tempurl%
\url{https://arxiv.org/abs/2508.05995}
\showURL{%
\tempurl}


\bibitem[Yu et~al\mbox{.}(2024)]%
        {yu2025look}
\bibfield{author}{\bibinfo{person}{Fei~Xu Yu}, \bibinfo{person}{Zuyuan Zhang}, \bibinfo{person}{Emily Grob}, \bibinfo{person}{Gina Adam}, \bibinfo{person}{Sean Coffey}, \bibinfo{person}{Nathaniel~D Bastian}, {and} \bibinfo{person}{Tian Lan}.} \bibinfo{year}{2024}\natexlab{}.
\newblock \showarticletitle{Look-ahead robust network optimization with generative state predictions}. In \bibinfo{booktitle}{\emph{AAAI 2025 Workshop on Artificial Intelligence for Wireless Communications and Networking (AI4WCN)}}.
\newblock


\bibitem[Zhao(2024)]%
        {zhao2024balf}
\bibfield{author}{\bibinfo{person}{Zhenjun Zhao}.} \bibinfo{year}{2024}\natexlab{}.
\newblock \showarticletitle{Balf: Simple and efficient blur aware local feature detector}. In \bibinfo{booktitle}{\emph{Proceedings of the IEEE/CVF Winter Conference on Applications of Computer Vision}}. \bibinfo{pages}{3362--3372}.
\newblock


\end{thebibliography}

\end{document}